\documentclass{article}
\usepackage{ijcai24}

\usepackage{times}
\usepackage{soul}
\usepackage{url}
\usepackage[hidelinks]{hyperref}
\usepackage[utf8]{inputenc}
\usepackage[small]{caption}
\usepackage{graphicx}
\usepackage{amsmath}
\usepackage{amsthm}
\usepackage{booktabs}
\usepackage{algorithm}
\usepackage{algorithmic}
\usepackage[switch]{lineno}
\usepackage{bbding}
\usepackage{booktabs}
\usepackage{nicematrix}

\title{A Survey on Robotics with Foundation Models: toward Embodied AI}

\author{
Zhiyuan Xu$^{*}$
\and
Kun Wu$^{*}$
\and
Junjie Wen
\and
Jinming Li
\\
Ning Liu
\and
Zhengping Che
\and
Jian Tang$^{\dagger}$\\
\affiliations
Midea Group\\
\emails
\{xuzy70, wukun12, wenjj25, lijm295, liuning22, chezp, tangjian22\}@midea.com
\thanks{~Equal contributions}
\thanks{~Corresponding author}
}

\usepackage{booktabs}                                   
\usepackage{multirow}                                   
\usepackage{tablefootnote}                              
\usepackage[symbol]{footmisc}                           
\usepackage{amsmath,amssymb}                            
\usepackage{xcolor}                                     
\usepackage{enumitem}                                   
\usepackage{subcaption}                                 
\usepackage{graphicx}
\usepackage{arydshln}
\usepackage{float}
\usepackage{wrapfig}

\newcommand{\eat}[1]{}                                  

\usepackage{amsmath}
\usepackage{amsthm}

\usepackage{bm}

\usepackage{amssymb}
\usepackage[T1]{fontenc}

\usepackage{pifont}

\allowdisplaybreaks

\begin{document}

\maketitle

\begin{abstract}
While the exploration for embodied AI has spanned multiple decades, it remains a persistent challenge to endow agents with human-level intelligence, including perception, learning, reasoning, decision-making, control, and generalization capabilities, so that they can perform general-purpose tasks in open, unstructured, and dynamic environments. Recent advances in computer vision, natural language processing, and multi-modality learning have shown that the foundation models have superhuman capabilities for specific tasks. They not only provide a solid cornerstone for integrating basic modules into embodied AI systems but also shed light on how to scale up robot learning from a methodological perspective. This survey aims to provide a comprehensive and up-to-date overview of foundation models in robotics, focusing on autonomous manipulation and encompassing high-level planning and low-level control. Moreover, we showcase their commonly used datasets, simulators, and benchmarks. Importantly, we emphasize the critical challenges intrinsic to this field and delineate potential avenues for future research, contributing to advancing the frontier of academic and industrial discourse.
\end{abstract}

\section{Introduction}

Although robots are extensively used in various fields like manufacturing, logistics, and specialized operations, they have been confined to controlled environments with pre-programmed actions. 
The advent of machine learning and deep learning has ushered in vision-guided planning and deep reinforcement learning-based control, reducing the need for precise environment or manipulation object modeling that is impractical in dynamic real-world scenarios. Nevertheless, due to training on task-specific data with limited model capacity, these methods still struggle with generalizability and transferability, impeding the deployment of robots into everyday environments with general-purpose tasks. 

Recently, foundation models have become pivotal in advancing embodied AI.
Pre-trained on massive, internet-scale datasets consisting of text, images, audio, and video, these large models, with billions of parameters, demonstrate remarkable capabilities. Examples include Large Language Models (LLMs)~\cite{openai2023gpt}, Large Vision Models (LVMs)~\cite{kirillov2023sam}, and large multi-modal Vision-Language Models (VLMs)~\cite{radford2021learning}, presenting exceptional understanding, reasoning, interaction, and generative abilities.
However, these models are not yet able to interact directly with entities in the physical world, i.e., they merely provide humans and agents with auxiliary information and are not the decision-makers, so current applications are limited to the Internet.
Therefore, how to integrate foundation models into decision-making models harmoniously is a key question for embodied AI.


\begin{figure*}[t]
    \centering
    \includegraphics[width=1.0\textwidth]{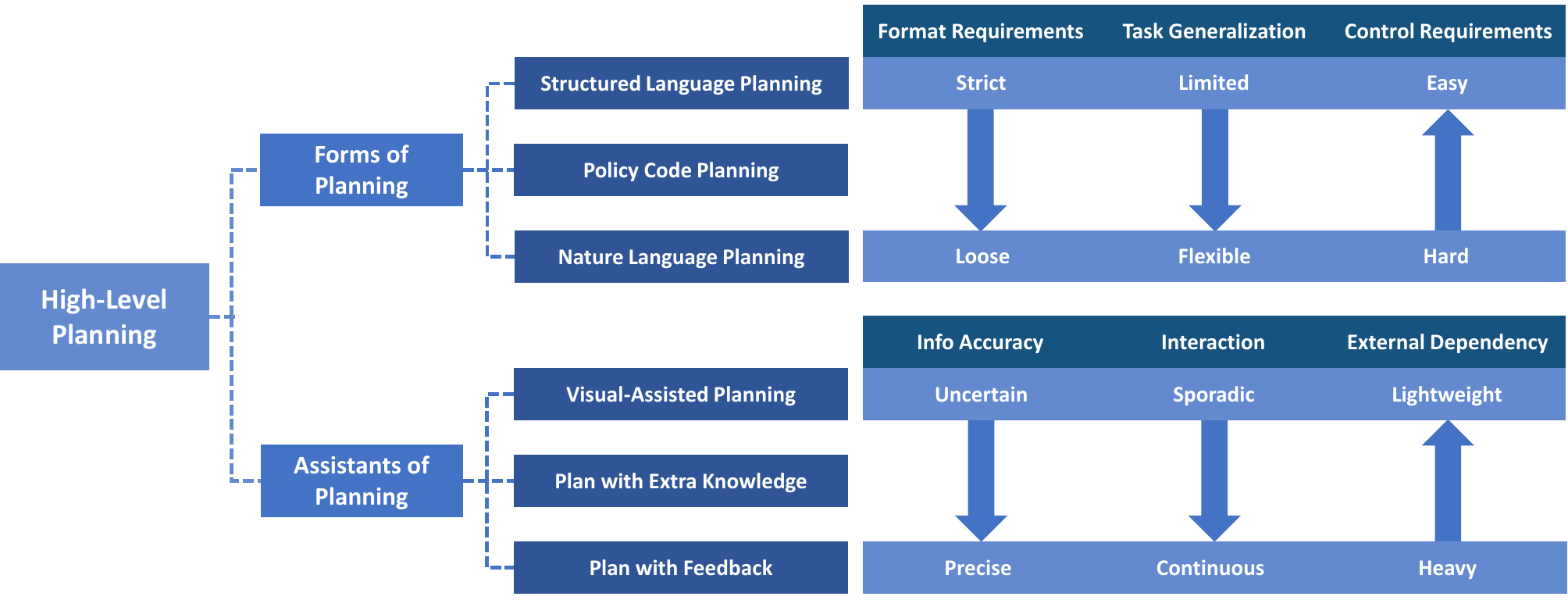}
    \caption{Taxonomy of foundation models for high-level planning and their properties.}
    \label{fig:tax}
\vspace{-10pt}
\end{figure*}

In this survey, we methodically review recent papers that apply foundation models to address challenges in autonomous robotic manipulation, categorizing them into two fundamental ways: 
1) \textbf{High-Level Planning:} This involves processing natural language commands from users and observations from interactive environments, culminating in step-by-step decision-making. Foundation models aid in interpreting commands, perceiving environments, and reasoning through complex tasks using their underlying world knowledge; 
2) \textbf{Low-Level Control:} The focus is on implementing execution commands while monitoring environmental and robotic states, ultimately determining precise execution parameters, such as the position of the robot arm end-effector or the degree of joints. The capacity of foundation models enables learning multiple tasks and making fine-grained execution decisions simultaneously.
Furthermore, recent advancements merge these two aspects, training large robotic foundation models to handle both planning and control. Their synergies, like the collaboration of the cerebrum and cerebellum, significantly enhance the reasoning and execution ability to handle complex and long-horizon tasks in a few-shot or zero-shot manner.

Despite recent demonstrations showcasing significant progress toward achieving general-purpose tasks, there remains a considerable journey ahead before robots can be widely deployed in everyday scenarios. In this survey, we also investigate the commonly used datasets, simulators, and benchmarks in this research field, enabling readers to quickly grasp the subject and apply their knowledge with existing resources. More importantly, we emphasize some persisting challenges and propose several potential research directions. 



\section{Foundation Models for High-Level Planning}


As the complexity of long-horizon tasks increases, training a single policy model to complete an entire task based on simple and concise instruction remains challenging. The model needs not only to acquire multiple skills but also to understand the sequential execution order among these skills, making the training process inherently intricate. Considering the demonstrated prowess of vision and language foundation models in intricate reasoning and contextual generalization, it is a natural progression for the robotics community to explore the application of foundation models in addressing these challenges. 
From the perspectives of forms (i.e., Section~\ref{sec:planform}) and assistants (i.e., Section~\ref{sec:planassis}) of planning, we review how to apply foundation models for understanding user command and planning long-horizon tasks. The taxonomy of foundation models for high-level planning is illustrated in Fig.~\ref{fig:tax}.

\subsection{Forms of Planning}\label{sec:planform}
Foundation models, such as LLMs and VLMs, are capable of generating a range of planning formats, spanning from highly structured forms like Planning Domain Definition Language (PDDL) to expressive programming code and even to the more flexible natural language. 
Stricter formats offer greater precision and are more easily interpreted by robotic controllers, yet their predefined nature restricts the expressiveness of foundation models. 
This limitation means the models are confined to generating only within these structures, potentially reducing their efficacy in zero-shot tasks. 
Conversely, a more loosely structured format, while offering increased expressiveness and better suitability for interpreting user commands in natural language, may pose execution challenges for robotics. 
For instance, natural language planning steps necessitate a sophisticated low-level controller for effective understanding and execution. 


PDDL is a typical family of structured languages that defines a robotic planning problem. Its standardization has the benefit of making research more reusable and easily comparable, though at the cost of some expressive power, compared to domain-specific systems. 
LLMs, like GPT-4~\cite{openai2023gpt}, first translate planning problems into PDDL descriptions, then are parsed by a standard PDDL planner. Finally, parsing results will be further translated to finite state automata~\cite{mpmsdm2023}, executable programming code~\cite{isrllm2023}, or natural lanugage~\cite{llm+p2023}.

\subsubsection{Policy Code Planning}
Policy code planning decouples natural language instructions into several steps in code form. Notably, the planning form is a well-organized and order-sensitive program that involves intricate planning. 
Code excels at providing an accurate description of the spatial position and exclusive reference of objects. Moreover, employing code as a high-level plan offers the advantage of expressing functions or feedback loops, processing perception outputs, and parameterizing control primitive APIs.
Code as Policies (CaP)~\cite{liang2023code} first presents a robotic-centric formulation of language model generated programs (LMPs) executed on real systems. 
With sophisticated prompt engineering, CaP could establish a target program according to the natural language instruction and recursively define new functions to generalize to novel tasks. 
Furthermore, ProgPrompt~\cite{singh2023progprompt} encodes the world state into natural language and then embeds it in prompts for LLMs to generate policy code.
Instruct2Act~\cite{huang2023instruct2act} utilizes LLMs to map multi-modal instructions to sequential code for robotic manipulation tasks.
OEIEA~\cite{sarch2023oeiea} generates policy code by human-robot dialogue that offers more detailed information and utilizes the VLMs to get scene descriptions.

\subsubsection{Natural Language Planning}
Natural language generation is the inherent ability of foundation models, which possess good interpretability. However, natural language may lack clear expression that misleads agents and aborts tasks. 
To mitigate this, CAPEAM~\cite{capeam2023} utilizes a sub-goal planner to extract sub-goals and concrete objects from task instruction, thereby reducing expressive ambiguity through the composition of these sub-goals and objects. Moreover, it leverages external models to evaluate sub-goals to verify their feasibility, correctness, and clarity. 
ViLA~\cite{vila2023} directly applies the most advanced multi-modal foundation models, GPT4-V, to generate nature language planning for long-horizon tasks through chain-of-thought reasoning and rich web-scale knowledge. 
Furthermore, RSFT~\cite{zhu2024rsft} proposes to identify fast and slow thinking for more precise planning and execution, which refers to straightforward actions like picking and multi-step reasoning tasks like rearranging, respectively. 
\subsection{Assistants of Planning}\label{sec:planassis}
The complexity of robot planning extends beyond mere language input or output. 
Consequently, researchers are exploring diverse methods to augment foundation models, like LLMs, with additional tools for more effective planning. 
This includes integrating vision models into LLMs to assist in target identification and incorporating external feedback to guide the robotic manipulation. What's more, combining extra common sense knowledge with foundation models is beneficial for reasoning and planning as well. These enhancements aim to bridge the gap between linguistic processing and practical perceptual information in robotics.

\subsubsection{Visual-Assisted Plan}
Language-based Foundation models, such as LLMs~\cite{touvron2023llama}, often require the support of conventional deep learning models (e.g., 2D/3D detection models) or other large vision-language models~\cite{llava,zhu2024llava} to process image data for task planning.
Inner Monologue~\cite{huang2022inner} directly integrates detection results of vision foundation models into the prompt of LLM for planning or re-planning. 
PHYSOBJECTS~\cite{physobjects2023} utilizes LLMs to generate initial plans and grounds them with the help of VLMs by querying physical concepts (e.g., material, fragility) of everyday objects. 
Vision foundation models can also serve as value or affordance functions to assist LLMs in next-step planning.
Specifically, SayCan~\cite{saycan2023} introduces a vision model to predict the probability of successfully executing said skills in a plan. This model collaborates with the LLMs, which evaluates the probability of selecting each skill, to jointly determine the next step in the process. 
However, the skills are predefined in SayCan, which lacks task generalizability. 
To mitigate this, GD~\cite{gd2023} directly predicts the probability of each token for the next step, aided by a vision model. The selection of the next token then involves a joint decoding process, which combines the probabilities estimated by both the LLMs and the vision model.

VLMs can directly integrate LLMs and visual information for task planning, which excels at grounding language into the environments and understanding tasks. 
A simple and efficient way to utilize VLMs is to substitute LLMs with VLMs in the origin framework. 
ViLA~\cite{vila2023} utilizes the novel GPT4-V model to generate plans showcasing strong zero-shot reasoning and generalization capability.  
PaLM-E~\cite{driess2023palm} jointly trains an embodied multi-modal vision-language model by blending the robotic data and routine vision-language data, which performs well both at embodied reasoning tasks and vision-language tasks. 
Though VLMs significantly improve the grounding capability, they fall short in fine-grained semantic understanding, which is essential for tiny object manipulation and dexterous operation. OCI-Robotics~\cite{wen2023oci} fine-tunes VLMs to predict the precise position of target objects, alleviating this deficiency to some extent. 

\subsubsection{Plan with Extra Knowledge}
Foundation models typically lacks domain-specific knowledge, hence it may cause errors in task planning in some scenarios e.g. make a cup of coffee with coffee machine. 
This extra knowledge can be acquired from self-planning, i.e., ``inside'', and other models, i.e., ``outside''. 
ClIN~\cite{majumder2023clin} learns from its planning failures, recursively using these insights for subsequent planning until success. 
CAPEAM~\cite{capeam2023} introduces an environment-aware memory system that records the historical locations of objects to enhance agent interactions. 
LMZSP~\cite{lmzsp2023} translates each step of the origin task plan from LLMs to precise admissible actions through another pre-trained foundation model.
PLAN~\cite{plan2023} constructs commonsense-infused prompts from external knowledge, which serves as an external prompt for enhanced task planning.

\subsubsection{Plan with Feedback}
The aforementioned methods, like PaLM-E or ViLA, employ preliminary, one-time planning, whereas interactive planning dynamically evolves throughout task execution. Precisely, the foundation models predict subsequent actions, assisted by instructions and real-time feedback, continually updating their plans following each step. 
Typical feedback consists of: 
1) feedback from the environment, indicating the signals from task completion or the success of actions. 
Inner Monologue~\cite{huang2022inner} incorporates closed-loop language feedback based on the success signals of actions into robot planning, significantly enhancing the completion of high-level instructions;
2) feedback from humans, involving direct human-robot interaction and serving as an intuitive signal to enhance planning capability. 
IRAP~\cite{irap2023} uses a question-and-answer format to obtain essential task-relevant knowledge, like the specific location of a target, facilitating precise robot instructions generation.

\begin{table*}[t]\scriptsize
    \caption{Taxonomy of foundation models for low-level control and their properties.}
    \label{tab:low_level_policy_overview}
    \centering
    \resizebox{\textwidth}{!}{
        \begin{NiceTabular}{c|c|ccc|cccc}
        \toprule
        &  & \multicolumn{3}{c}{\bf Minimum Requirements} & \multicolumn{4}{c}{\bf Properties} \\
        {\bf Directions} & {\bf Category} & {\bf Reward} & {\bf Optimal Data} & {\bf Suboptimal Data} & {\bf Causality} & {\bf Sample Efficiency} & {\bf End-to-End} & {\bf Generalization} \\
        \midrule
        \multirow{2}*{\shortstack{Policy \\ Learning}} & RL & \ding{51} &  & & \ding{51} & Low & \ding{51} & Low \\
        & IL &  & \ding{51} & & \ding{51} & High & \ding{51} & Low \\
        \midrule
        \multirow{2}*{\shortstack{Environment \\ Modeling}} & Forward Dynamic &  &  & \ding{51} & \ding{51} & Low & & Improve \\
        & Inverse Dynamic &  &  & \ding{51} &  & High & & Improve \\
        \midrule
        \multirow{2}*{\shortstack{Representation \\ Learning}} & Frozen Models &  & \ding{51} &  &  & High & & Improve \\
        & Learned Models &  & \ding{51} & \ding{51} & \ding{51} & Low & \ding{51} & Improve \\
        \bottomrule
        \end{NiceTabular}
    }
\end{table*}

\section{Foundation Models for Low-Level Control}

In addition to using the decomposition and planning capabilities of large foundation models to help robots do high-level planning for complex long-horizon tasks, numerous studies also propose leveraging the recognition, reasoning, and generalization capabilities of the language and vision foundation models for generating accurate low-level control.
Moreover, employing internet-scale data for training the foundation models also paves a promising way for training robotic foundation models.
In this section, we focus on the works dedicated to the robotic low-level control, i.e., the ultimate goal is to learn a policy to output actions like target joint position, target 6 DoF poses, and so on.
We categorized these methods into three classes from the perspective of different focused components in the learning process:
1) Policy learning.
2) Environment modeling.
3) Representation learning.
As shown in Table~\ref{tab:low_level_policy_overview}, we provide a taxonomy of foundation models for low-level control and their properties.

\subsection{Policy Learning}


\subsubsection{Reinforcement Learning}
RL problems are typically formulated using the Markov Decision Process (MDP).
The goal is to maximize the total reward of the tasks
by interacting with the environments and learning in a trial-and-error manner.
One of the major barriers preventing RL from being used in real-world applications is reward shaping~\cite{ng1999policy} 
that RL requires carefully designed reward functions to learn. 
Therefore, many recent studies
have turned their attention to utilizing the knowledge and understanding of various tasks contained in large foundation models to provide rewards and reduce laborious human involvement.
After observing that textual task instructions typically express the outcome of robot behavior rather than the details of the low-level actions, 
L2R~\cite{yu2023language} proposes the use of reward functions as a bridge between language instructions and low-level actions and explores the use of the code generation capability of LLMs to convert task descriptions into reward functions.  
Using the generated reward, the trajectory optimization and reinforcement learning algorithms can learn a good policy. 
Eventually, L2R performs well on robotic arms for dexterous manipulation tasks.
Another important challenge lies in the low sample efficiency of RL methods.
Recently, many studies
have argued that the foundation models can greatly accelerate the development of the above techniques by infusing the learning process with more prior knowledge learned from internet-scale data.
Specifically, FRL~\cite{ye2023foundation} presents a foundation reinforcement learning framework and instantiates an actor-critic method, FAC, which leverages the value, reward, and policy prior knowledge from foundation models
to achieve sample-efficient learning, robustness to noisy priors, and minimal human intervention.
Moreover, the success of using ballooned internet-scale data with transformer architecture in the vision and language domain has fueled the imagination about how embodied intelligence can evolve.
For instance, Q-Transformer~\cite{chebotar2023q} discusses design decisions for using transformer architecture to learn Q-function and multi-task policy given large-scale offline datasets.
\subsubsection{Imitation Learning}
Considering the MDP setting, where the difference from RL is that the expert demonstrations are given and the reward function is unknown, the goal of imitation learning is to recover the expert policy and thus achieve higher rewards or task success rates.
Since the distribution of expert demonstrations is typically narrow, the major problem is that the compounding error exists and the generalization ability of learned policy is poor.
Inspired by the fact that LLMs and VLMs can achieve strong zero-shot performance in the open-set world via training with the overwhelming amount of data,
an intuitive and straightforward way is to extend this mode into embodied AI and use large-scale robotic data to train a policy with high capability.
Many studies have put the above idea into practice, the best known of which would be RT-1~\cite{brohan2022rt}.
In RT-1, the authors spent 17 months collecting massive amounts of data on robot manipulation from 13 robots, including 130k trajectories for 700 tasks.
Equipped with a high-capacity transformer-based architecture and pre-trained vision and language models, RT-1 can not only achieve high success rates on the training tasks but also perform better than baseline methods in new test tasks and scenarios, showing better generalization ability.

\subsection{Environment Modeling}

Besides learning a policy directly from demonstrations or interactions with environments, it is also important to learn the environment (world) model, which can be used in conjunction with other algorithms, such as reinforcement learning and trajectory optimization, to generate better policies.

\subsubsection{Forward Dynamic Learning}

Forward dynamic learning predicts the next state based on the current or historical observations and actions.
This learning objective incorporates the nature of causality because the next state is the result of an action made at the current state. 
Thus, it can effectively model the environment, containing the physical rules.
The trade-off is that it is difficult and requires a large amount of data to build an accurate dynamic model.
As a seminal work, SWIM~\cite{mendonca2023structured} learns world models with forward dynamics from internet-scale human play videos.
In order to overcome the morphological differences between humans and robots, the method learns in a high-level abstract and structured action space, i.e., predicting the target poses and waypoints.
Subsequently, the model can be fine-tuned using a small portion of the robotic data in an unsupervised way, which does not require a reward function as well as task-related supervision.

\subsubsection{Inverse Dynamic Learning}

Inverse dynamic learning aims to predict the action given the current and next states as input.
It is non-causal since it leverages information from the future to recover past behaviors.
Thus, this learning method reduces the difficulty of training and is not as data-hungry as forward dynamic learning.
~\cite{brandfonbrener2023inverse} builds extensive experiments on robot manipulation tasks over various domains.
The empirical results show that 1) learning inverse dynamic can perform better than other per-training objectives when learning from scratch, and 2) it can extract smoother and more discriminative representations with higher sample efficiency compared to forward dynamic objective.


\subsection{Representation Learning}

Representation, one of the most fundamental concepts in the era of deep learning, has a pivotal impact on the final performance of various tasks.
Currently, vision and language foundation models trained with large internet-scale data can extract beneficial representations for images and text instructions, providing a solid cornerstone for downstream embodied AI tasks.
Also, they are able to align with the features of the text.
For example, the foundation models can recognize areas of apples in an image based on the word "apples" in the instructions.
Following this spirit, many studies
take LLMs and VLMs as fixed representation extractors and establish a harmonious composition and pipeline.

\subsubsection{Frozen Foundation Models}

One of the most straightforward approaches is to use the large model as a read-standing component for some specific functions, such as object detection~\cite{liu2023grounding}, semantic segmentation~\cite{kirillov2023sam}, and object tracking~\cite{yang2023track}.
For instance, SMS~\cite{sharma2023semantic} presents a plug-and-play semantic module that provides more detailed and comprehensive semantic information for downstream embodied AI tasks. 
Through the use of an open vocabulary object detection model, it is able to recognize objects in the current scene and, in conjunction with the LLMs, explicitly builds a semantic occupancy distribution as the feature of the scene.

\subsubsection{Learned Control Models}

To extract descriptive and intelligible representations from robotic data, many works
propose to use the techniques in different fields like unsupervised learning~\cite{alloghani2020systematic}, self-supervised learning~\cite{balestriero2023cookbook}, and transfer learning~\cite{zhuang2020comprehensive}.
More specifically,
Motivated by the pattern of human acting and learning in the real world, 
DualMind~\cite{wei2023imitation} proposes a similar two-stage learning strategy for learning a generalist policy that can handle multiple decision-making tasks. 
In the first phase, the policy is trained in a self-supervised manner to learn the underlying common knowledge in the world. 
Subsequently, in the second phase, a small part of the parameters is used to train a conditional policy based on multi-modal prompts. 
The method uses an Encoder-Decoder Transformer architecture combined with Token-Leaner and a cross-attention mechanism for enhancing feature extraction ability.
\section{Datasets, Simulators, and Benchmarks}

\subsection{Datasets}
\label{sec: datasets}


The development of foundation models in the domains of CV and NLP over recent years owes much of its success to the utilization of large-scale datasets.
Analogously, for the substantial improvement of embodied AI capabilities, the imperative lies in scaling up the robotic datasets, which serves as a requisite pathway to facilitate significant advancement in embodied AI.
Compared to the vision and language dataset that emphasizes abstract, 2D, and web-based information, the datasets for embodied AI focus more on 3D information in real-world environments, the affordance of objects, and the interactions between the robots and objects, which are based on the constraints of real-world physical laws. We divide the datasets for embodied AI into three categories, including object, human, and robotic datasets.

\subsubsection{Object Datasets}
Object datasets usually contain a large number of synthesized or real scanned items and their corresponding labels, including categories, meshes, point clouds, etc.
These datasets can be used for various downstream tasks, including robotic environment building, perception, novel view synthesis, object reconstruction, object generation, and robotic manipulation tasks.
For example, OmniObject3D~\cite{wu2023omniobject3d} scans from the real world to collect a total of 6,000 everyday objects of 190 types and aligns the category with commonly used 2D datasets like ImageNet.
It also provides rich object annotations, including multi-view images and even videos.
Moreover, all objects are scanned using professional scanners and thus have accurate shapes and high-fidelity appearance, providing high-quality environments and scenes.

\subsubsection{Human Datasets}
Human datasets are usually in the form of videos and contain people performing activities from the first- or third-person view in different environments and scenarios, from doing chores in the kitchen to biking outdoors.
These data are very broad and relatively easy to collect, e.g., downloading human activity videos from YouTube.
They contain real-world information about the appearance of the objects and environments, the human-object interactions, and physical rules embedded in the movement and dynamics, which are beneficial to the embodied AI if this knowledge can be successfully transferred to the agents.
In the latest work Ego-Exo4D~\cite{grauman2023ego}, the authors build a massive multi-view human video dataset containing a total of 1,422 hours of video divided into 5,625 instances, each between 1 and 42 minutes, with more than 800 participants from 13 different cities around the world.

\subsubsection{Robotic Datasets}
Robotic datasets contain demonstrations of directly controlling robots to perform tasks in the real world or a simulator, with various collection methods like teleoperation devices, pre-trained agents, and so on.
However, these collection processes are time-consuming and labor-intensive, e.g., setting up the environment in which the robots perform the tasks and teaching operators to use the relevant equipment.
This leads to the fact that it is very difficult to scale the robotic datasets to very large sizes, which is one of the challenges embodied AI faces.
RT-X~\cite{padalkar2023open} is leading the way and taking an important step forward in this regard.
It assemble a dataset from 22 different robots collected through a collaboration between 21 institutions, demonstrating 527 skills (160,266 tasks).
Based on such large-scale data, the trained agent shows positive performance in zero-shot generalization and cross-morphology transfer learning.



\subsection{Simulators}
In the field of embodied AI, high-fidelity simulators are crucial for efficient training and reducing the gap between simulation and real-world application. Key challenges in simulator development include 1) Real-time soft-body material simulation, where advanced simulators like ManiSkill2 use the material point method (MPM) for more realistic and complex interactions compared to traditional methods like finite element method (FEM); 2) Support for multiple controllers to accommodate various tasks and control methods, with simulators like ManiSkill2~\cite{gu2023maniskill2} and Nvidia Isaac Sim offering versatility in action spaces for tasks ranging from obstacle avoidance to pick-and-place; 3) The ability to handle the parallel computation of multiple environments, essential for large-scale data collection and training in reinforcement learning, where simulators like ManiSkill2 and Isaac Gym demonstrate superior computational efficiency and functionality in managing numerous simultaneous environments.

\subsection{Benchmarks}

In embodied AI, benchmarks are crucial for assessing diverse systems and algorithms to ensure fair comparisons. However, the vast scope of embodied AI, encompassing a wide range of tasks, systems, and environments, poses a challenge in creating a universal benchmark. Two notable benchmarks are FMB \cite{luo2024fmb}, which offers a benchmark for testing and improving robotic manipulation skills with tasks like grabbing, moving, and assembling objects using 3D-printed items for easy replication and study, and ManiSkill2~\cite{gu2023maniskill2}, presenting a collection of 20 targeted operational tasks designed to address common challenges in operational skill benchmarks. Both provide comprehensive platforms for developing and evaluating embodied AI agents, catering to the field's complex and varied requirements.
\section{Discussions and Future Directions}

The exploration of robotics integrated with foundation models is currently in its nascent phase. Despite significant advancements, numerous challenges impede the practical application of robots in real-world environments. In this section, we highlight some existing challenges and potential future avenues of research in this area. 

\subsection{Synergies of Planning and Control}

Current embodied AI has achieved unprecedented success on simple tasks such as grasping and placing objects. 
However, for long-horizon tasks, such as object rearrangement, learning a single policy is very difficult because errors accumulate quickly with steps.
Therefore, it is important to learn high-level planning and low-level control simultaneously and make them collaborate seamlessly.
Recently, RT-2~\cite{zitkovich2023rt} and RT-X~\cite{padalkar2023open} present significant advancements for this objective.
As a milestone in robotics work, RT-X absorbs its predecessors, RT-1 and RT-2, including the use of VLMs, LLMs, 
and transformer-based policy networks for learning. 
More importantly, authors from 21 institutions have collected massive amounts of data on robot manipulation, including 527 skills and 160,266 tasks from 22 robots with different morphologies.
Using these data, RT-X demonstrates a strong ability to transfer among different morphologies and generalize to new tasks and scenarios.

Compared to integrating planning and control in a single large policy network, many previous works
study how to decompose complex tasks rationally and learn skills effectively.
Nevertheless, they use only pre-defined primitive tasks (skills) and can not generalize to new tasks, which is a prerequisite for robots to be successfully deployed in real-world applications.
At present, the development of vision and language foundation models also provides a strong boost to this direction, for example, using LLMs for more rational planning and expanding the skill library.
More specifically,
given the high-dimensional visual observation history as the input, Skill Transformer~\cite{huang2023skill} predicts both high-level skills and low-level actions concurrently.
It consists of a skill prediction module and a transformer-based policy that can model the long-horizon sequences and dependencies without making every subtask completely isolated. 


\subsection{Hallucination of Foundation Model}

Foundation models have achieved remarkable success due to their ability to learn from massive datasets and generalize to domains beyond their training scope. However, significant concerns have been raised regarding the safety and reliability of these models, particularly in relation to their tendency to 'hallucinate' or generate misleading information~\cite{ouyang2022training}. A case in point is GPT-4V, which can sometimes interpret images by including objects that do not actually exist. This issue becomes critically important in fields like robotics, where inaccurate interpretations or instructions provided by foundation models can lead to serious malfunctions or failures. Additionally, the aspect of model safety is vital in the development of embodied agents~\cite{liu2023query,perez2022red}, where the model is integrated with a physical entity. Misuse of these models can lead to catastrophic outcomes, reminiscent of scenarios depicted in science fiction novels. In these applications, the physical capabilities of the agents, combined with potentially flawed decision-making processes, underscore the need for rigorous safety protocols and reliable operational frameworks to prevent adverse outcomes~\cite{brunke2022safe}. 

\subsection{Efficient Data Collection}

A key observation gained from the fact that large vision and language foundation models show super generalization ability as well as emergent abilities~\cite{wei2022emergent} is that scaling the size of datasets is perhaps one of the most important things to do.
The current scale of the robotic dataset is nowhere near the magnitude used for the vision and language foundation models.
For example, the vision and language dataset LAION-5B~\cite{schuhmann2022laion} contains 5.75 billion text-image pairs, while the dataset in VIMA~\cite{jiang2023vima} only collects 650k successful demonstrations.
The collection of robotics data poses an even more formidable challenge compared to acquiring purely visual and language data.
A fraught example is that RT-1~\cite{brohan2022rt} spends 17 months to collect 130k episodes.
With this collecting speed, it is essentially impossible to collect robotic data of a magnitude comparable to the vision and language internet-scale dataset.
How to collect robotic data more efficiently remains a key and open question.


One promising solution is to utilize a high-fidelity simulator for large-scale trajectory collection or training and then deploy the agent to real-world applications via Sim2Real methods~\cite{huang2023went}.
However, there is still a distribution shift between simulators and the real world, both in terms of external appearance and internal dynamic, which can significantly degrade the performance, and the Sim2Real approach cannot fully compensate for this part of the gap.

Another solution is to utilize the extra human activity datasets~\cite{grauman2022ego4d}, as we discuss in Section~\ref{sec: datasets}.
These data are relatively easy to collect by downloading them from the internet, and they contain real-world observations and dynamic information.
Experimental results from the work~\cite{huo2023human} demonstrate that incorporating extra data can enhance performance.
However, it is difficult to align this information with the inputs and outputs of robots since robots and humans have very different morphologies.

\subsection{Augmenting Existing Datasets}

Compared to vision and language tasks, collecting large-scale training data for embodied AI tasks is much more difficult and costly.
People need time-consuming training to become proficient in the use of teleoperation equipment and are susceptible to a variety of interfering factors that can lead to mistakes, such as fatigue and distracting events.
Therefore, how to effectively augment the existing robotic datasets 
is an unavoidable question.

The straightforward approach is to enhance the image data from the perspective of computer vision, which uses commonly used data augmentation techniques~\cite{xu2023comprehensive}, 
such as modifying brightness, flipping, cropping, noise injection, etc. 
However, these methods may destroy the semantic information in the images, which degrades the final performance. 
Recently, data augmentation methods without destroying the semantic information have been proposed to enhance the generalization performance of the policy further by utilizing VLMs and LLMs. 
One of these encouraging works, ROSIE~\cite{yu2023scaling}, reformulates the data augmentation problem as an image inpainting problem. 
It uses the VLMs to delineate the task-relevant region of the images and uses the Imagen Editor, a state-of-the-art diffusion model, to augment the region based on textual instructions. 
Finally, the robot manipulation policy trained on the augmented data shows enhanced generalization capabilities on hundreds of tasks.
However, these methods essentially do not increase the number of robotic trajectories, meaning that when the agent enters a dangerous or out-of-distribution (OOD) state, it still does not know how to recover from the error.

\subsection{Efficiency on Computation and Deployment}
 
The effective integration of foundation models significantly enhances robot learning, such as PaLM-E-562B~\cite{driess2023palm} and RT-1~\cite{brohan2022rt}. 
%
However, both the training and inference phases demand substantial computational resources.
Training the huge models on large-scale datasets necessitates a bunch of high-end GPUs.
Therefore, efficient pre-training and fine-tuning techniques like LoRA~\cite{hu2021lora} and MiniGPT-4~\cite{zhu2023minigpt} are essential to mitigate training costs. 
Moreover, deploying these large models on resource-constrained edge devices presents challenges because of concerns over user data privacy and the need for real-time performance in robotic applications, which makes cloud deployment impractical. 
%
%
Consequently, optimizing the computation efficiency of large models through model compression techniques and designing lightweight models is vital for enabling deployment on edge devices~\cite{dey2023cerebras}.

\subsection{Safety and Interpretation of Robot}

In the deployment of robotic systems within real-world scenarios, particularly in areas with a high concentration of people and objects, such as human-robot collaborations or indoor environments, ensuring running safety is imperative. 
This involves mitigating the risk of collisions and preventing injuries to people and property damage to objects. 
While certain approaches~\cite{zhao2021model} offer theoretical guarantees for robot systems through the utilization of Lyapunov functions, they only consider ideal and simple experimental environments with only low-dimensional inputs and necessitate precise modeling of the environment.
However, these real-world applications with high-dimensional inputs like images are more intricate, e.g., many assumptions will not hold anymore, so how to ensure security in real-world environments poses a formidable challenge.
In contrast, alternative methods~\cite{farid2022task} focus on using out-of-distribution (OOD) detection to facilitate real-time identification of errors made by the robot. 
However, these methods grapple with the challenge of generalization in novel scenarios and the formulation of recovery strategies following errors, constituting a crucial area for ongoing research.


\section{Conclusion}

In this paper, we offer an in-depth overview of the latest applications of foundation models in embodied AI, with a special emphasis on autonomous robotic manipulation. We methodically categorize existing literature based on high-level planning strategies and low-level control mechanisms. Furthermore, this review includes an extensive survey of the robots, simulators, and benchmarks that are most commonly utilized in experimental research in this area. It concludes with a discussion of the key challenges and outlines promising research directions. Through this survey, we hope to advocate for a concerted effort from both academic and industrial spheres to advance foundation models in robotics toward a new era of generalized, embodied AI.

\clearpage

\bibliographystyle{named}
\bibliography{ijcai24}

\end{document}